\documentclass[runningheads]{llncs}
\usepackage{amsmath}

\usepackage{amssymb}
\usepackage{amsmath}
\usepackage{latexsym}
\usepackage[ruled,vlined]{algorithm2e}
\usepackage{etoolbox}  

\makeatletter
\patchcmd{\algorithmic}{\addtolength{\ALC@tlm}{\leftmargin} }{\addtolength{\ALC@tlm}{\leftmargin}}{}{}
\makeatother
\usepackage{url}
\usepackage{xcolor}

\usepackage{hyperref}
\usepackage{cite}
\usepackage{graphicx}
\newcommand{\etal}{\textit{et al.}}

\begin{document}
\title{Attention for Image Registration (AiR): an unsupervised Transformer approach}
\titlerunning{Transformer based Image Registration}
%
%
\authorrunning{Zihao WANG et al.}

\author{Zihao~Wang\inst{1,2}\and
Herv\'{e}~Delingette\inst{1}
}

%
\institute{Inria Sophia-Antipolis, Epione Team, Valbonne, France \\
\email{zihao.wang@inria.fr}
\and Universit\'{e} C\^{o}te d'Azur, Nice, France \\
}
\maketitle              
\begin{abstract}
Image registration is a crucial task in signal processing, but it often encounters issues with stability and efficiency. Non-learning registration approaches rely on optimizing similarity metrics between fixed and moving images, which can be expensive in terms of time and space complexity. This problem can be exacerbated when the images are large or there are significant deformations between them.
Recently, deep learning, specifically convolutional neural network (CNN)-based methods, have been explored as an effective solution to the weaknesses of non-learning approaches. To further advance learning approaches in image registration, we introduce an attention mechanism in the deformable image registration problem.
Our proposed approach is based on a Transformer framework called AiR, which can be efficiently trained on GPGPU devices. We treat the image registration problem as a language translation task and use the Transformer to learn the deformation field. The method learns an unsupervised generated deformation map and is tested on two benchmark datasets.
In summary, our approach shows promising effectiveness in addressing stability and efficiency issues in image registration tasks. The source code of AiR is available on Github.\footnote{\url{https://github.com/MatheoZihaoWang/AIR_Transformer_Image_Registration}}

\end{abstract}
\section{Introduction}
Image registration is a widely researched topic in areas such as remote sensing, radar engineering, and medical imaging, where the matching of two images for further processing, such as image synthesis, stitching, or segmentation, is a fundamental task \cite{INTRO_DL_REMOTE, INTRO_REVIEW, INTRO_RADAR, WANG2021101990}. Image registration can be divided into rigid and non-rigid registration, with deformable image registration being one of the most challenging branches. This is because finding the deformation field between images to be matched is a highly nonlinear problem. Traditional approaches rely on collecting image features and using various similarity metrics to measure the matching quality between image pairs during optimization \cite{Intro_Reg_review}.


Here's an edited version of your paragraph:

Deep learning has been widely adopted in medical image analysis, with various convolutional neural network (CNN) variants introduced for image registration, producing remarkable results in many research works. Wu \etal \cite{Intro_DL_Wu} proposed a CNN for automatic feature extraction to replace manual feature extraction during registration. Cao \etal \cite{Intro_DL_CAO} introduced a meticulously designed CNN architecture to learn the non-linear mapping between input images and deformation fields. The training dataset was prepared using a diffeomorphic demons-based registration dataset. De Vos \etal \cite{Intro_DL_de_Vos} used an unsupervised Deep Learning Image Registration (DLIR) framework that uses CNN to regress the mapping between image pairs and deformation maps. This framework is unsupervised, as the CNN's output is used directly to warp the moving images, generating deformed images with transposed convolutions that enable backpropagation for resampling operations. Krebs \etal \cite{Intro_DL_Krebs} proposed learning a probabilistic deformation model based on variational inference of a conditional variational auto-encoder (CVAE), which maps the deformation representation in a latent space. The trained CVAE can be used to generate deformations for an unseen image.

Since the proposal of Transformers-based learning models, they have rapidly expanded from the field of natural language processing (NLP) to the entire machine learning community \cite{INTRO_transformer_NIPS, INTRO_transformer_CV, INTRO_transformer_CV2}. The Transformer has already conquered the NLP regime as a very successful application of the attention concept, almost replacing the traditional widely-used RNN/LSTM models. The recent advance of Transformer applications in image/video analysis implies that Transformers have the potential to show their power in structural data processing. The Transformer is a different learning model that does not require convolutional kernels for feature representation but learns data inherent relationships through the \textbf{attention mechanism} \cite{METHOD_TRANS}. Recent works in the computer vision community report that Transformer-based deep learning methods have achieved state-of-the-art performance on many datasets.

\begin{figure*}[t!]
    \centering
    \includegraphics[width=\columnwidth]{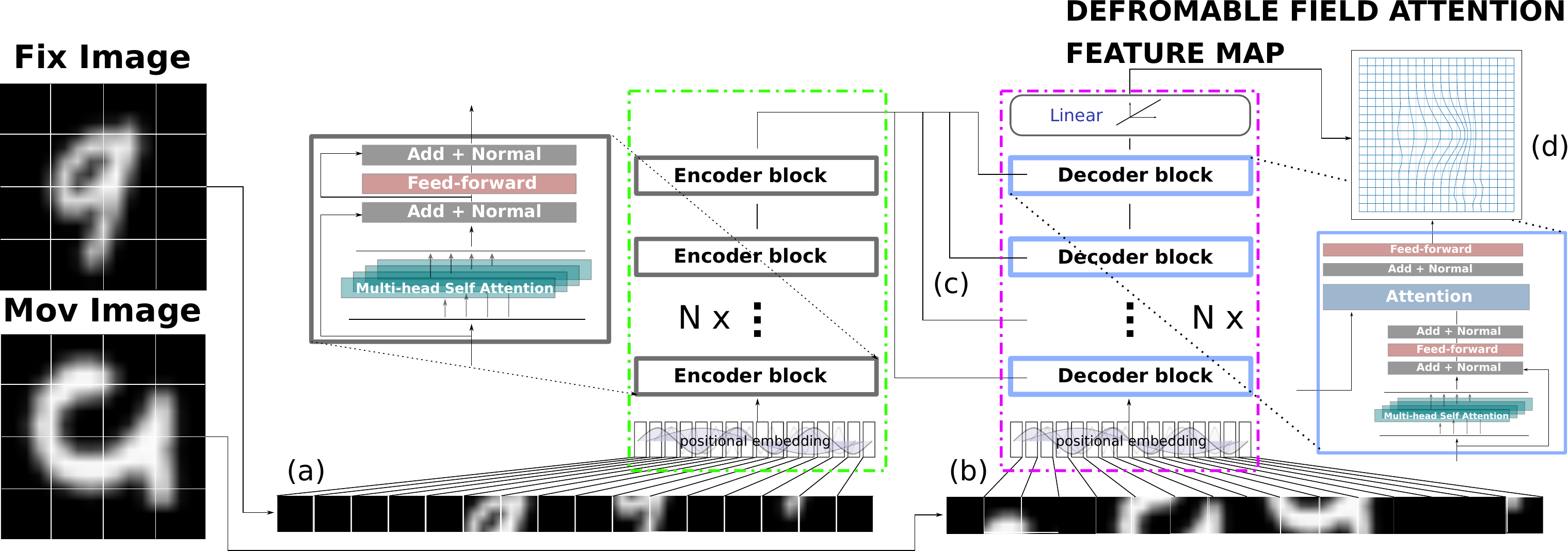}
    \caption{Transformer framework used for deformation attention feature map prediction. The proposed Transformer consists of encoder and decoder modules. The encoder module (shown in the green dotted line box) takes the fixed image patches as input and learns the representation of memory attention features using the self-attention mechanism. The decoder module (shown in the purple dotted line box) takes the attention features of the fixed image from the encoder (memory) and the self-attention features of the moving image as inputs to predict the deformable features that can transform the moving image into a fixed image.}
    \label{fig:transformer}
\end{figure*}

One representative work that introduces Transformers to computer vision is the vision Transformer (ViT) proposed by \cite{Intro_TRANS_Dosovitskiy} for image classification. The ViT takes divided patches $p_i$ of a given image $I$ as inputs (see Fig.~\ref{fig:transformer} step (a)) and uses a Transformer as the attention feature extractor to generate features for classification. Another work that uses a Transformer for object detection is DETR \cite{Intro_TRANS_Carion}. DETR accepts the image and detection objects for the Transformer as inputs and generates corresponding bounding box information of the target objects on the images. Recently, the Swin Transformer \cite{Intro_TRANS_Swin} achieved state-of-the-art performance for multi-task learning. The Swin Transformer uses a similar hierarchical structure to build different attention perception field sizes on target images.

In this paper, we propose a novel unsupervised image registration framework called AiR, which is based on the attention mechanism. Our approach introduces the Transformer model to the field of image registration, different from previous CNN-based methods. The framework uses the Transformer to learn the deformation information between fixed and moving images without convolution operations. Unlike many prior works in the computer vision community, the proposed framework does not rely on any neural network backbone as a prior feature extractor.

To the best of our knowledge, the proposed framework is the first Transformer-based image registration method. The proposed Transformer-based image registration framework is unsupervised, and training deformation fields are not necessary. Additionally, we propose a multi-scale attention parallel Transformer framework to better solve the incompetent cross-level feature representation of conventional Transformers.

\section{Method}
\subsection{Attention for Image Registration (AiR)}
The proposed attention-based image registration framework is based on the Transformer model \cite{METHOD_TRANS}. The Transformer model can be viewed as a dimensional isotropic projection:
\begin{equation}
T_{\theta}: T(x) \rightarrow z; x, z \in \mathbb{R}^{n\times d}
\end{equation}
where $T$ is the projection, parameterized by a set of parameters $\theta: {W_q,W_k,W_h,W_c,W_r}$. The computational pipeline start from the input tensor $x$ to output tensor $z$ is given through:
\begin{equation}
\begin{aligned}
Query = W_q \cdot x; Key = W_k \cdot x; Value = W_v \cdot x; \\
\alpha = softmax(<Query, Key>/\sqrt{k}); \\
z' = \sum_{h=1}^{H}{W_c}\sum_{j=1}^{J}{\alpha \cdot {Value}};\\
z = NormRelu(z')
\end{aligned}
\end{equation}
In our problem setting, we aim for the Transformer to learn the deformation map between two images, where the prediction target tensor $z$ represents the corresponding deformation map for warping the moving image to the fixed image. Unlike the usual CNN-based methods that merge the fixed and moving images in the channel dimension as input to regress the relationship between the deformation maps and input image pairs, we treat the image registration problem as an image translation problem similar to using the Transformer in NLP tasks.

Specifically, we want the Transformer to learn the deformation map (attention features) that can translate the moving image (foreign language) into the fixed image (native language). Based on this idea, we introduce an encoder-decoder Transformer framework for image registration, using the encoder to model the fixed image and the decoder to model the moving image. The self-attention maps that come from the fixed image are injected into the attention blocks of the decoder. The final output attention map of the Transformer is used to warp the moving image to match the fixed image.


To feed the images to the Transformer, the fixed $I_f$ and moving $I_m$ image pairs are first divided into $i \cdot i$ patches (see Fig.~\ref{fig:transformer}, step (a), where $i = 4$) and then embedded by linear projection with added position embedding to provide positional information, following the approach used in \cite{Intro_TRANS_Dosovitskiy}.

The fixed image patches are then fed to the Transformer encoder network, which encodes the attention features of the fixed images. The Transformer encoder network contains $N$ recursive blocks, as shown in the green box of Fig.\ref{fig:transformer}). The output of the Transformer decoder network is processed with a linear projection layer to generate the deformation attention feature maps.

\begin{figure*}[ht!]
    \centering
    \includegraphics[width=\columnwidth]{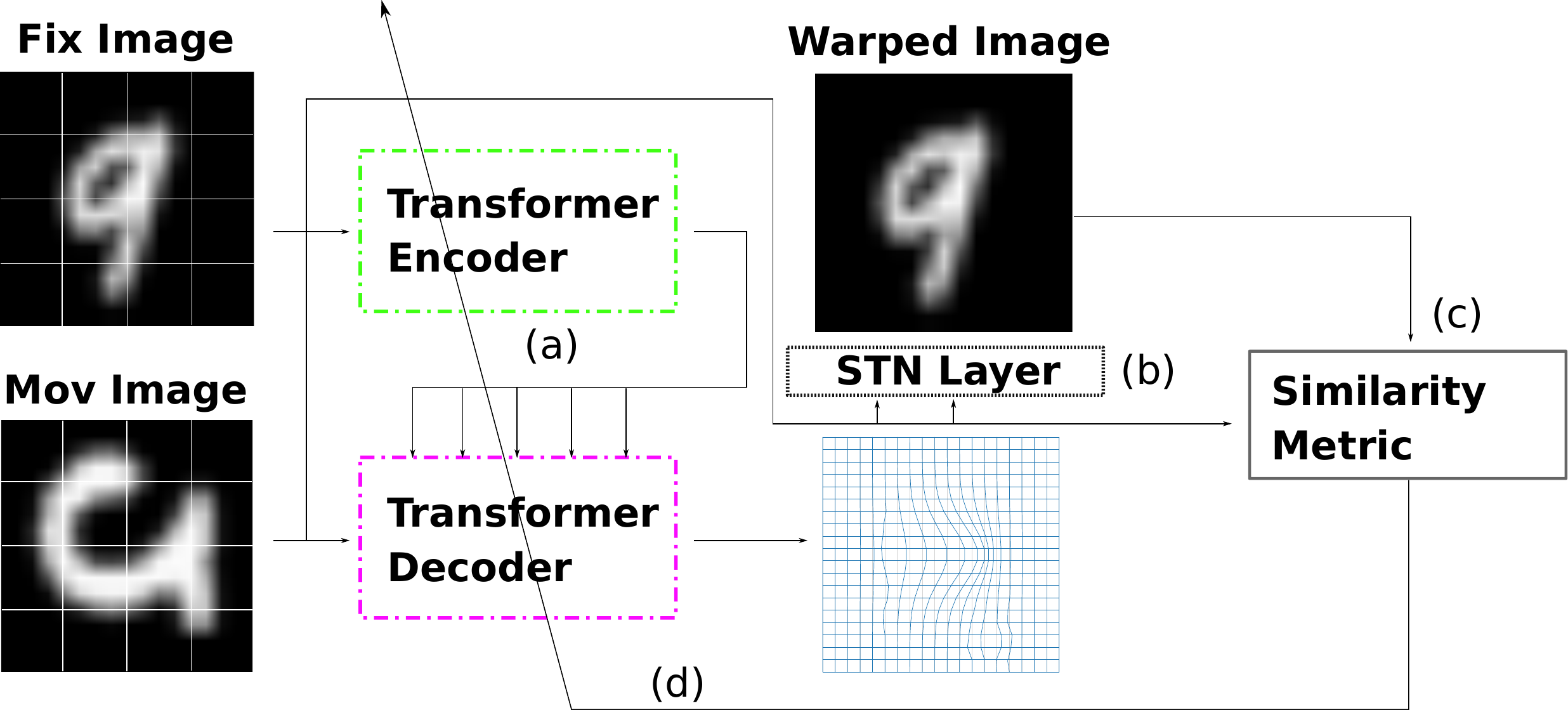}
    \caption{The Transformer based image registration framework. In step (a): The input images are converted to deformation feature maps by the Transformer. In step (b): Those feature maps are converted to deformation field for a sampler to warp the moving image. Steps (c) -(d): The warped moving image is quantified by the similarity metric for computing the gradients flow for back-propagation.}
    \label{fig:framework}
\end{figure*}

Similar to DLIR \cite{Intro_DL_de_Vos}, the generated deformation features are then processed by a spatial Transformer (STN) layer \cite{METHOD_SNT} for generating the displacement field to sample the moving images. As shown in Fig.~\ref{fig:framework}, the output of the STN layer is then used for sampling the moving image to generate the warped image. Finally, backpropagation is achieved by optimizing the similarity metric that measures the similarity between the deformed images and the fixed images.

\section{Multi-scale Parallel Attention Transformer}
CNN-based methods can use hierarchical feature layer superposition and different sizes of convolutional kernels for feature expression at different levels. However, as the Transformer is not a convolutional-based method, its ability for cross-feature level feature extraction is comparatively weaker. This problem was also observed in our experimental section for a single-level Transformer (patch size equal to 2) based AiR framework.

To achieve multi-level feature learning for the Transformer, we propose a Multi-scale Attention Parallel Transformer (MAPT) that can learn features from different perception scales. The MAPT consists of $N$ Transformers ($N$ decoders and $N$ encoders). For each Transformer, they adapt to different sizes of patches as inputs and generate $N$ different attention feature maps $F_N$. The $N$ feature maps are then sampled to a uniform size, and then added together with normalized weighting ratios to get the final deformable feature map $F$.

\section{Experiments and Evaluation}
\subsection{Dataset and Experiment Details}
We evaluated the single-scale and multi-scale Transformer-based AiR frameworks in comparison with CNN-based DIRNets on the MNIST dataset and Fashion-MNIST \cite{lecun2010mnist, fasionmnist}. During the training period, randomly selected image pairs were used to feed the AiR transformer for gradient descent. We used $20\%$ of the data for testing the performance of the AiR Transformer and $80\%$ for training. The Transformer in our experiment contained 1 block ($N=1$) for both the Transformer encoder and decoder networks. The dropout ratio of the encoder and decoder layers was set to 0.5. The attention layers of the Transformer consisted of 4 heads for both the multi-head self-attention and attention layers. The projection dimension was set to 16 for both the encoder and decoder networks. We trained all the frameworks with Adam optimization using a learning rate of $lr=0.5e-3$. All experiments were run on an Asus ESC8000 GPU node with two Xeon SP Gold 5115 @ 2.4 GHz CPU and using one GeForce GTX 1080 Ti GPU. The program was implemented with the PyTorch framework.
\subsection{Results and Evaluation}
\begin{figure*}[ht!]
    \centering
    \includegraphics[width=\columnwidth]{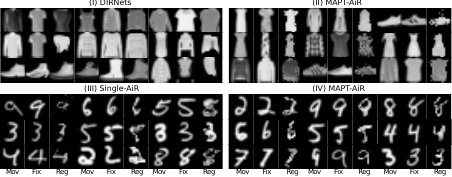}
    \caption{Registration results of different approaches on different datasets: (I) CNN based method performance on Fashion-MNIST dataset. (II) Multi-scale Parallel Attention Transformer (MPAT) based AiR results on Fashion-MNIST dataset. (III) Single Transformer based AiR on MNIST dataset. (IV) MPAT based AiR results on the MNIST dataset.}
    \label{fig:res}
\end{figure*}

Figure~\ref{fig:res} depicts the registration outcomes of different approaches. The generated results demonstrate that all the frameworks are capable of learning the warping mapping between the moving and fixed images. However, the different algorithms exhibit varying deformation qualities.

In the upper row of Figure~\ref{fig:res}, we utilize DIRNets, a CNN-based registration framework, to register the randomly paired data sampled from the Fashion-MNIST dataset. The deformation maps of the DIRNets results are sometimes over or under deformed. On the other hand, the MAPT-AiR framework can capture globally shaped deformations as the target images' shapes are well-matched. However, the MAPT-AiR still lacks the detailed portrayal compared to the CNN-based method. The lower row of Figure~\ref{fig:res} demonstrates the registration performance comparison between Single-AiR and MAPT-AiR. Both the algorithms exhibit reasonable performances. Nonetheless, the Single-AiR framework may fail to match large deformations, for example, the number 4 and 9. In contrast, the MAPT-AiR performs better in modeling large deformations, as seen in the cases of numbers 9, 7, and 8.

Overall, the figure illustrates that the transformer-based registration methods, while producing sampled images with similar overall shape to the fixed images, suffer from geometric detail defects and noise in the warped images. This defect may be attributed to the quality of the generated deformable fields. Recent works suggest that transformers are inferior to CNN in terms of pixel-level details processing ability \cite{RES_DEPTH} \cite{RES_TRANS_DETAILS}. This problem is also observed in our experiment.

\begin{table}
\centering
\caption{Quantitative comparison between different methods on MNIST handwriting dataset.}
\label{tab:quantative}
\begin{tabular}{clccc}
                                  & \multicolumn{1}{c}{\textbf{Original Image}} & \textbf{DIRNets} & \begin{tabular}[c]{@{}c@{}}\textbf{AiR-Single }\\\textbf{Attention}\end{tabular} & \begin{tabular}[c]{@{}c@{}}\textbf{AiR-Parallel}\\\textbf{Attention}\end{tabular}  \\ 
\hline
\textbf{MSE}                      & 0.0647$\pm$0.029                            & 0.033$\pm$0.014  & 0.038$\pm$0.015                                                                  & \textbf{0.027$\pm$0.015 }                                                          \\
\textbf{PSNR}                     & 12.941$\pm$2.218                            & 16.966$\pm$2.897 & 15.65$\pm$2.638                                                                  & \textbf{17.311$\pm$2.522}                                                                     \\
\textbf{Smooth DICE }             & 0.756$\pm$0.039                             & 0.809$\pm$0.033  & 0.797$\pm$0.032                                                                  & \textbf{0.827$\pm$0.034}                                                           \\
\hline
\end{tabular}
\end{table}
We conducted a quantitative evaluation of the different frameworks using three metrics, which are presented in Table \ref{tab:quantative}. The results show that the multi-scale attention-based Transformer framework achieves the best performance across all three metrics. On the other hand, using a single attention Transformer is not better than the CNN-based registration framework. This is expected since the single-scale Transformer cannot learn features that cross different scale levels, while CNN is powerful in modeling such features. However, as shown in Table \ref{tab:quantative}, this limitation can be overcome by using the multi-scale attention framework.
\section{Conclusion}
In this paper, we introduced a novel Attention-based Image Registration (AiR) framework built on the Transformer model, which is distinct from the commonly used CNN-based approaches. We demonstrated in our experiments that the proposed framework can achieve the goal of unsupervised deformable image registration and that the Transformer model can deliver a plausible performance for this task. However, as the Transformer is a relatively new research topic in computer vision, there are still some limitations for its widespread application compared to CNN-based methods. One challenge is the Transformer's GPU-bound nature, which prohibits its use on conventional computing systems. There is still much work to be done to address the computational complexity of the Transformer framework.

\bibliographystyle{unsrt}
\bibliography{refs}
\end{document}